\documentclass[sigconf]{acmart-me}

\usepackage{booktabs} 
\usepackage{url}
\usepackage{color}
\usepackage{enumitem}
\usepackage{multirow}
\usepackage{subfloat}
\usepackage{subfig}
\usepackage{balance}
\setcopyright{rightsretained}
\usepackage[printonlyused]{acronym}

\acmDOI{}

\acmISBN{}

\acmConference[MediaEval'19]{Multimedia Evaluation Workshop}{27-29 October 2019}{Sophia Antipolis, France \\ Github: https://github.com/vlbthambawita/MedicoTask\_2019\_paper\_2} 
\acmYear{}
\copyrightyear{}

\acmPrice{}

\newacro{MLP}  [MLP]  {multi-layer perceptron}
\newacro{MSE} [MSE] {mean square error}
\newacro{MAE} [MAE] {mean absolute error}
\newacro{CNN} [CNN] {convolutional neural networks}

\begin{document}
\title{Extracting temporal features into a spatial domain using \\autoencoders for sperm video analysis
\vspace{-8pt}}

\author{Vajira Thambawita\textsuperscript{1,2}, P{\aa}l Halvorsen\textsuperscript{1,2}, Hugo Hammer\textsuperscript{1,2}, Michael Riegler\textsuperscript{1,3}, Trine B. Haugen\textsuperscript{2}}
\affiliation{
\textsuperscript{1}SimulaMet, Norway \ \ \ {}
\textsuperscript{2}Oslo Metropolitan University, Norway\ \ \ \
\textsuperscript{3}Kristiania University College, Norway \\
}
\email{Contact: vajira@simula.no}

%
%
%
%
%

\renewcommand{\shortauthors}{Thambawita et al.}
\renewcommand{\shorttitle}{2019 Medico Medical Multimedia}

\begin{abstract}
In this paper, we present a two-step deep learning method that is used to predict sperm motility and morphology based on video recordings of human spermatozoa. First, we use an autoencoder to extract temporal features from a given semen video and plot these into image-space, which we call feature-images. Second, these feature-images are used to perform transfer learning to predict the motility and morphology values of human sperm. The presented method shows it's capability to extract temporal information into spatial domain feature-images which can be used with traditional convolutional neural networks. Furthermore, the accuracy of the predicted motility of a given semen sample shows that a deep learning-based model can capture the temporal information of microscopic recordings of human semen.
\end{abstract}

%
%
%
%
%


\maketitle

\section{Introduction}
\label{sec:intro}
The 2019 Medico task~\cite{medico2019overview} focuses on automatically predicting semen quality based on video recordings of human spermatozoa. This is change from previous years which have mainly focused on image classification of images taken from the gastrointestinal tract~\cite{pogorelov2018medico, riegler2017multimedia}. For this year's task, we look at predicting the morphology and motility of a given semen sample. Motility is defined by three variables, namely, the percentage of progressive, non-progressive, and immotile sperm. Morphology is determined by the percentage of sperm with tail defects, midpiece defects, and head defects. The organizers have provided a dataset consisting of 85 videos of different semen samples and a preliminary analysis of each, which is used as the ground truth. For this competition, the organizers have provided a predefined three-fold split of the VISEM dataset~\cite{visem}, which contains 85 videos from different participants and a preliminary analysis of each semen sample. In the dataset paper, the authors presented baseline \ac{MAE} values for motility and morphology. Furthermore, the importance of computer-aided sperm analysis can be identified from the previous works which have been done over the last few decades~\cite{mortimer2015future, Urbano2017, Dewan2018}.

To solve this year's task, we propose a deep learning-based method consisting of two steps - (i) unsupervised feature extraction using an autoencoder~\cite{autoencoders} and (ii) video regression using a standard \ac{CNN} and transfer learning.  
The autoencoder we use is different from the state-of-the-art autoencoders used to extract video features~\cite{chong2017abnormal, yang2015unsupervised} as they use autoencoders to extract feature vectors which are used with long-short memory models or \ac{MLP}s. In contrast, we use autoencoders to extract feature-images for use in CNNs.



\section{Approach}
\label{sec:approach}

\begin{figure*}[t!]
    \centering
    \includegraphics[scale=0.85]{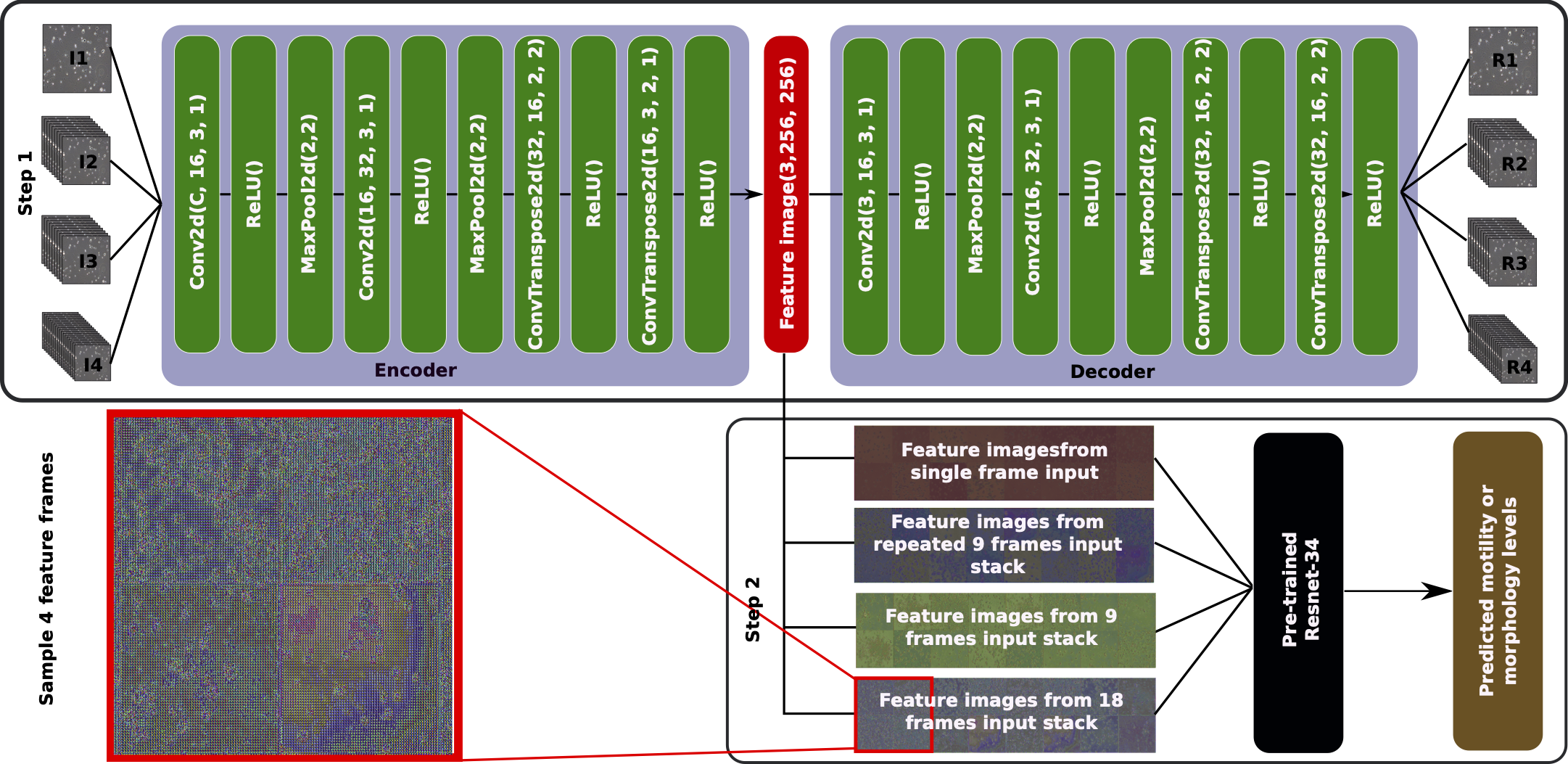}
    \caption{A big picture overview of our two step deep learning model: Step 1 - an autoencoder architecture used to extract image features, Step 2 - the pre-trained Resnet-34 \ac{CNN} for predicting the regression values of motility and morphology, I1, I2, I3 and I4 - input frames extracted from the video dataset, R1, R2, R3 and R4 - reconstructed data corresponding to the input data I1, I2 I3 and I4, sample 4 feature frames shows extracted 4 feature images from the autoencoder after training 2000 epochs (actual resolution of a feature image is 256X256 which is equal to the original frame size of the input data)}
    \label{fig:model_architecture}
\vspace{-10pt}
\end{figure*}

Our method can primarily be split into two distinct steps. First, we use an autoencoder to extract temporal features from multiple frames of a video into a feature-image. Second, we pass the extracted feature-image into a standard pre-trained \ac{CNN} to predict the motility and morphology of the spermatozoa in a given video. In this paper, we present the preliminary results for four experiments based on four different input types. The first input type (I1) uses a single raw frame. Input type two (I2) is a stack of identical frames copied across the channel-dimension. The third (I3) and fourth (I4) input type stack 9 and 18 consecutive frames from a video respectively. 

The first two experiments (using I1 and I2) were performed as baseline experiments. The two other experiments (using I3 and I4) were performed to see how the temporal information affects the prediction performance of the approach. For all input types, we split the extracted datasets into three folds based on the folds provided by the organizers. Then, three-fold cross-validation was conducted to evaluate our four experiments. An overview of all experiments is shown in Figure~\ref{fig:model_architecture}.
\vspace{-7pt}

\subsection{Step 1 - Unsupervised temporal feature extraction}
\label{sec:input_data}


In step 1, we trained an autoencoder that takes an input frame or frames (I1, I2, I3 or I4) from the sperm videos as depicted in Figure~\ref{fig:model_architecture}. Then, the encoder of the autoencoder extracted feature-images and passed them through the decoder architecture to reconstruct the input frame or frames back (R1, R2, R3, and R4). These extracted feature-images are different from traditional feature extractions of autoencoders because the traditional autoencoders extract feature vectors instead of feature-images. In this autoencoder, the \ac{MSE} loss function is used to calculate the difference between input data and reconstructed data. Then, this error value is backpropagated to train the autoencoder. After training 2,000 epochs, we use the encoder architecture of the autoencoder model to step 2. 
\vspace{-7pt}

\subsection{Step 2 - \ac{CNN} regression model}
We have selected the pre-trained ResNet-34 \cite{resnet_34} as our basic \ac{CNN} to predict the values of motility and morphology of the sperm videos. However, any pre-trained \ac{CNN} could be chosen for this step and in future work we will test and compare different ones in more detail. 
Firstly, we take an input frame or frames (I1, I2, I3 or I4) and pass through the pre-trained encoder model (only the encoder section of the autoencoder model) which was trained also from the same data inputs in an unsupervised way. Then, the outputs of the encoder model were passed through the \ac{CNN} model which has a modified last layer to output three prediction values for motility or morphology. 
\section{Results and Analysis}
\label{sec:results_and_analysis}

According to the average \ac{MAE} values shown in Table~\ref{tbl:results},  the average motility values of input I3 and I4 shows the best results among other motility values of input I1 and I2. These performance improvements imply that our model is able to learn temporal features into a spatial feature image representation. Furthermore, input I4 which uses 18 stacked frames shows the best motility average values compared to input I3. This performance gain shows that to predict the sperm motility in sperm videos, it is better to analyze more frames at the same time. This might be due to the fact that the behaviour of sperm is something that needs to be observed over time and not in single frames. Moreover, the predictions for our base case inputs I1 and I2 show the same average values. This shows that our model learns temporal information from different sperm video frames. Otherwise, it would be shown different average values for our two base case inputs I1 and I2. 

When we consider the predicted morphology average in Table~\ref{tbl:results}, it shows values that are almost equal to each other.  This is expected because the morphology of a sperm is something that can be observed using a single frame. In contrast to predicting accurate morphology, the predicted morphology values support the prove that our model has the capability to learn temporal data from multiple frames because motility predictions show an improvement when we increase the number of frames analyzed simultaneously.


\begin{table}[]
\vspace{10pt}
\caption{Mean absolute error values collected from the proposed method from different inputs: I1, I2, I3 and I4}
\vspace{-10pt}

\begin{tabular}{llrrrr}
\toprule
&  & \multicolumn{2}{c}{Motility} & \multicolumn{2}{c}{Morphology} \\ 
Input & Fold & \multicolumn{1}{c}{MAE} & Average & \multicolumn{1}{c}{MAE} & \multicolumn{1}{c}{Average} \\ \midrule

\multirow{3}{*}{I1} 
  & Fold 1 & 13.330 & \multicolumn{1}{l}{\multirow{3}{*}{13.017}} & 5.698 & \multirow{3}{*}{5.715} \\
  & Fold 2 & 12.880 &                       \multicolumn{1}{l}{} & 5.748 &  \\
  & Fold 3 & 12.840 &                       \multicolumn{1}{l}{} & 5.698 &  \\ \midrule
 
\multirow{3}{*}{I2} 
  & Fold 1 & 12.890 & \multicolumn{1}{l}{\multirow{3}{*}{13.017}} & 5.573 & \multirow{3}{*}{\textbf{5.606}} \\
  & Fold 2 & 13.010 &                       \multicolumn{1}{l}{} & 5.593 &  \\
  & Fold 3 & 13.150 &                       \multicolumn{1}{l}{} & 5.653 &  \\ \midrule
 
\multirow{3}{*}{I3} 
  & Fold 1 & 10.850 & \multicolumn{1}{l}{\multirow{3}{*}{10.970}} & 5.567 & \multirow{3}{*}{5.632} \\
  & Fold 2 & 11.310 &                       \multicolumn{1}{l}{} & 5.748 &  \\
  & Fold 3 & 10.750 &                       \multicolumn{1}{l}{} & 5.580 &  \\ \midrule
 
\multirow{3}{*}{I4} 
  & Fold 1 & 9.462 & \multicolumn{1}{l}{\multirow{3}{*}{\textbf{9.427}}} & 5.900 & \multirow{3}{*}{5.777} \\
  & Fold 2 & 9.426 &  &                    5.738 &  \\
  & Fold 3 & 9.393 &  &                    5.692 &  \\ 
 
\bottomrule
\end{tabular}
\label{tbl:results}
\vspace{-10pt}
\end{table}


%
%

\section{Conclusion and Future works}
\label{sec:conclution}
In this paper, we proposed a novel method to extract temporal features from videos to create feature-images, which can be used to train traditional \ac{CNN} models. Furthermore, we show that the feature-images capture temporal present in a sequence of frames, which can be used to predict the motility of the sperm videos. 

This method can be improved by using different error functions to force the model to learn more temporal data. For example, researchers can experiment with variational autoencoders~\cite{kingma2013auto} and generative adversarial learning methods~\cite{goodfellow2014generative} to improve this technique. Additionally, it may be beneficial to embed long short-term memory units to investigate how our feature-images compare to actual extracted temporal features.

\newpage
\bibliographystyle{ACM-Reference-Format}
\def\bibfont{\small} 
\balance
\bibliography{sigproc} 

\end{document}